\documentclass[conference]{IEEEtran}
\usepackage{subcaption}

\usepackage{soul}
\usepackage{color}

\ifCLASSINFOpdf
  \usepackage[pdftex]{graphicx}
\else
   \usepackage[dvips]{graphicx}
\fi

\usepackage{algorithm}
\usepackage{algpseudocode}

\usepackage{blindtext}
\let\labelindent\relax 
\usepackage{enumitem}

\begin{document}
%
\title{Smart Augmentation\\ Learning an Optimal Data Augmentation Strategy}

\author{
\IEEEauthorblockN{Joseph Lemley}
\IEEEauthorblockA{Collage of Engineering and Informatics\\
National University of Ireland Galway\\
Galway Ireland\\
Email: j.lemley2@nuigalway.ie}
\and
\IEEEauthorblockN{Shabab Bazrafkan}
\IEEEauthorblockA{Collage of Engineering and Informatics\\
National University of Ireland Galway\\
Galway Ireland\\
Email: s.bazrafkan1@nuigalway.ie}
\and
\IEEEauthorblockN{Peter Corcoran}
\IEEEauthorblockA{Collage of Engineering and Informatics\\
National University of Ireland Galway\\
Galway Ireland\\
Email: peter.corcoran@nuigalway.ie}
}

\maketitle

\begin{abstract}

A recurring problem faced when training neural networks is that there is typically not enough data to maximize the generalization capability of deep neural networks(DNN). There are many techniques to address this, including data augmentation, dropout, and transfer learning. In this paper, we introduce an additional method which we call Smart Augmentation and we show how to use it to increase the accuracy and reduce overfitting on a target network. Smart Augmentation works by creating a network that learns how to generate augmented data during the training process of a target network in a way that reduces that networks loss. This allows us to learn augmentations that minimize the error of that network. 

Smart Augmentation has shown the potential to increase accuracy by demonstrably significant measures on all datasets tested. In addition, it has shown potential to achieve similar or improved performance levels with significantly smaller network sizes in a number of tested cases.

\end{abstract}


%
\IEEEpeerreviewmaketitle

\section {Introduction}

In order to train a deep neural network, the first and probably most important task is to have access to enough labeled samples of data. Not having enough quality labeled data will generate overfitting, which means that the network is highly biased to the data it has seen in the training set and, therefore will not be able to generalize the learned model to any other samples. In \cite{Bazrafkan2017} there is a discussion about how much the diversity in training data and mixing different datasets can affect the model generalization. Mixing several datasets might be a good solution, but it is not always feasible due to lack of accessibility. 

One of the other approaches to solving this problem is using different regularization techniques. In recent years different regularization approaches have been proposed and successfully tested on deep neural network models. The drop-out technique \cite{srivastava2014dropout} and batch normalization  \cite{ioffe2015batch} are two well-known regularization methods used to avoid overfitting when training deep models. 

Another technique for addressing this problem is called augmentation. Data augmentation is the process of supplementing a dataset with similar data that is created from the information in that dataset. The use of augmentation in deep learning is ubiquitous, and when dealing with images, often includes the application of rotation, translation, blurring and other modifications to existing images that allow a network to better generalize \cite{simard2003best}. 

Augmentation serves as a type of regularization, reducing the chance of overfitting by extracting more general information from the database and passing it to the network. One can classify the augmentation methods into two different types. The first is unsupervised augmentation. In this type of augmentation, the data expansion task is done regardless of the label of the sample. For example adding a different kind of noise, rotating or flipping the data. These kinds of data augmentations are usually not difficult to implement. 

One of the most challenging kinds of data expansion is mixing different samples with the same label in feature space in order to generate a new sample with the same label. The generated sample has to be recognizable as a valid data sample, and also as a sample representative of that specific class. Since the label of the data is used to generate the new sample, this kind of augmentation this can be viewed as a type of supervised augmentation.

Many deep learning frameworks can generate augmented data. For example, Keras \cite{chollet2015keras} has a built in method to randomly flip, rotate, and scale images during training but not all of these methods will improve performance and should not be used ``blindly''. For example, on MNIST (The famous handwritten number dataset), if one adds rotation, the network will be unable to distinguish properly between handwritten ``6'' and ``9'' digits. Likewise, a system that uses deep learning to classify or interpret road signs may become incapable of discerning left and right arrows if the training set was augmented with by indiscriminate flipping of images. 

More sophisticated types of augmentation, such as selectively blending images or adding directional lighting rely on expert knowledge. Besides intuition and experience, there is no universal method that can determine if any specific augmentation strategy will improve results until after training. Since training deep neural nets is a time-consuming process, this means only a limited number of augmentation strategies will likely be attempted before deployment of a model. 

Blending several samples in the dataset in order to highlight their mutual information is not a trivial task in practice. Which samples should be mixed together how many of them and how they mixed is a big problem in data augmentation using blending techniques. 

Augmentation is typically performed by trial and error, and the types of augmentation performed are limited to the imagination, time, and experience of the researcher. Often, the choice of augmentation strategy can be more important than the type of network architecture used \cite{Goodfellow2016}.

Before Convolutional Neural Networks (CNN) became the norm for computer vision research, features were ``handcrafted''. Handcrafting features went out of style after it was shown that Convolutional Neural Networks could learn the best features for a given task \cite{lemley2017deep}. We suggest that since the CNN can generate the best features for some specific pattern recognition tasks, it might be able to give the best feature space in order to merge several samples in a specific class and generate a new sample with the same label. Our idea is to generate the merged data in a way that produces the best results for a specific target network through the intelligent blending of features between 2 or more samples.

\section {Related Work}

Manual augmentation techniques such as rotating, flipping and adding different kinds of noise to the data samples, are described in depth in \cite{simard2003best} and \cite{DBLP:journals/corr/ChatfieldSVZ14} which attempt to measure the performance gain given by specific augmentation techniques. They also provide a list of recommended data augmentation methods. 

In 2014, Srivastava et al. introduced the dropout technique \cite{srivastava2014dropout} aiming to reduce overfitting, especially in cases where there is not enough data. Dropout works by temporarily removing a unit (or artificial neuron) from the Artificial Neural Network and any connections to or from that unit. 

Konda et al. Showed that dropout can be used for data augmentation by "projecting the the dropout noise within a network back into the input space". \cite{konda2015dropout} 

Jaderberg et al. devised an image blending strategy as part of their paper "Synthetic Data and Artificial Neural Networks for Natural Scene Text Recognition" \cite{DBLP:journals/corr/JaderbergSVZ14}. They used what they call ``natural data blending'' where each of the image layers is blended with a randomly sampled crop of an image from a training dataset. They note a significant (+44\%) increase in accuracy using such synthetic images when image layers are blended together via a random process. 

Another related technique is training on adversarial examples. Goodfellow et al. note that, although augmentation is usually done with the goal of creating images that are as similar as possible to the natural images one expects in the testing set, this does not need to be the case. They further demonstrate that training with adversarial examples can increase the generalization capacity of a network, helping to expose and overcome flaws in the decision function \cite{goodfellow2014explaining}.

The use of Generative Adversarial Neural Networks \cite{goodfellow2014generative} is a very powerful unsupervised learning technique that uses a min-max strategy wherein a 'counterfeiter' network attempts to generate images that look enough like images within a dataset to 'fool' a second network while the second network learns to detect counterfeits. This process continues until the synthetic data is nearly indistinguishable from what one would expect real data to look like.  Generative Adversarial Neural Networks can also be used to generate images that augment datasets, as in the strategy employed by Shrivastav et al. \cite{shrivastava2016learning}

Another method of increasing the generalization capacity of a neural network is called ``transfer learning''. In transfer learning, we want to take knowledge learned from one network, and transfer it to another \cite{pan2010survey}. In the case of Convolutional Neural Networks, when used as a technique to reduce overfitting due to small datasets, it is common to use the trained weights from a large network that was trained for a specific task and to use it as a starting point for training the network to perform well on another task.  

Batch normalization, introduced in 2015, is another powerful technique. It was discovered upon the realization that normalization need not just be performed on the input layer, but can also be achieved on intermediate layers. \cite{ioffe2015batch} 

Like the above regularization methods, Smart Augmentation attempts to address the issue of limited training data to improve regularization and reduce overfitting. As with \cite{goodfellow2014explaining}, our method does not attempt to produce augmentations that appear ``natural''. Instead, our network learns to combine images in ways that improve regularization. Unlike \cite{simard2003best} and \cite{DBLP:journals/corr/ChatfieldSVZ14}, we do not address manual augmentation, nor does our network attempt to learn simple transformations. Unlike the approach of image blending in  \cite{DBLP:journals/corr/JaderbergSVZ14}, we do not arbitrarily or randomly blend images. Smart augmentation can be used in conjunction with other regularization techniques, including dropout and traditional augmentation. 

\section {Smart Augmentation}
Smart Augmentation is the process of learning suitable augmentations when training deep neural networks. 

The goal of Smart Augmentation is to learn the best augmentation strategy for a given class of input data. It does this by learning to merge two or more samples in one class. This merged sample is then used to train a target network. The loss of the target network is used to inform the augmenter at the same time. This has the result of generating more data for use by the target network.  This process often includes letting the network come up with unusual or unexpected but highly performant augmentation strategies.

\subsection{Training Strategy for Smart Augmentation}
During the training phase, we have two networks: Network A, which generates data; and network B, which is the network that will perform a desired task (such as classification). The main goal is to train network B to do some specific task while there are not enough representative samples in the given dataset.  To do so, we use another network A to generate new samples. 

This network accepts several inputs from the same class (the sample selection could be random, or it could use some form of clustering, either in the pixel space or in the feature space) and generates an output which approximates data from that class. 
This is done by minimizing the loss function LA which accepts out1 and image i as input. Where out1 is the output of network A and mage i is a selected sample from the same class as the input.  The only constraint on the network A is that the input and output of this network should be the same shape and type. For example, if N samples of a P-channel image are fed to network A, the output will be a single P-channel image. 

\subsection{The Generative Network A and Loss Function}

The loss function can be further parameterized by the inclusion of $\alpha$ and $\beta$  as $f(L_A, L_B; \alpha, \beta)$. In the experiments and results sections of this paper, we examine how these can impact final accuracy. 
\begin{figure*}[h!]
  \label{smartaugManyA}
  \includegraphics[width=\textwidth]{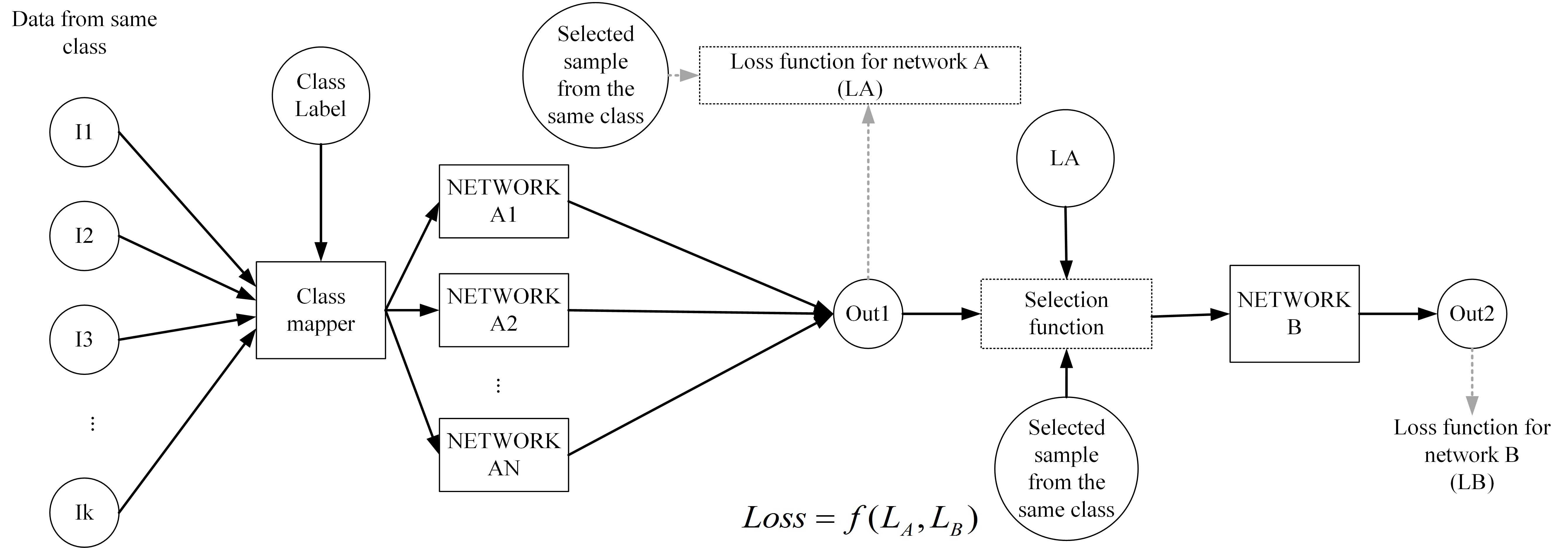}
  \caption{Smart augmentation with more than one network A}
\end{figure*}

Network A can either be implemented as a single network (figure 2) or as multiple networks, as in figure 1. Using more than one network A has the advantage that the networks can learn class-specific augmentations that may not be suitable for other classes, but which work well for the given class. 

\begin{figure*}[h!]
  \includegraphics[width=\textwidth]{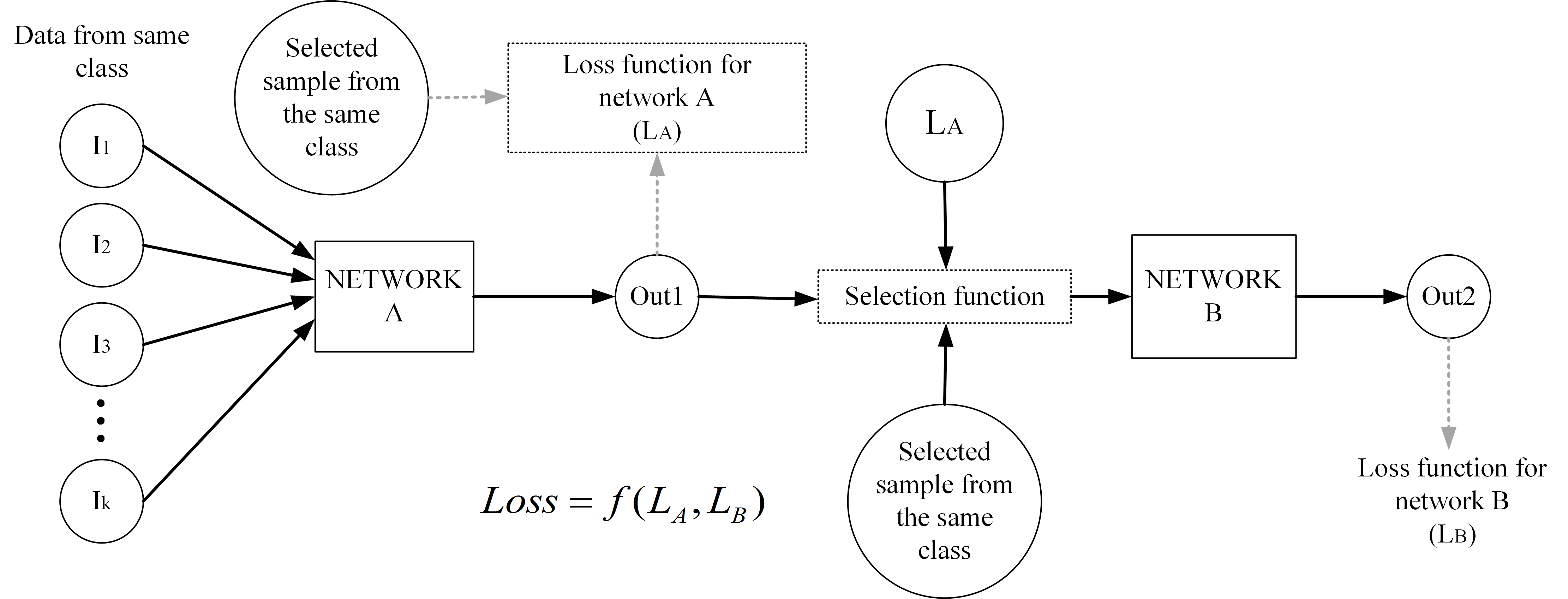}
  \caption{Diagram illustrating the reduced smart augmentation concept with just one network A}
\end{figure*}

Network A is a neural network, such as a generative model, with the difference that network A is being influenced by network B in the back propagation step, and network A accepts multiple samples as input simultaneously instead of just one at a time. This causes the data generated by network A to converge to the best choices to train network B for that specific task, and at the same time, it is controlled by loss function LA in a way that ensures that the outputs are similar to other members of its class. 

The overall loss function during training is $f(LA,LB)$ where $f$ is a function whose output is a transformation of LA and LB. This function could be an epoch-dependent function i.e. the function could change with the epoch number. 
In the training process, the error back-propagates from network B to network A. This tunes network A to generate the best augmentations for network B. After training is finished, Network A is cut out of the model and network B is used in the test process.
The joint information between data samples is exploited to both reduce overfitting and to increase the accuracy of the target network during training. 

\subsection{How Smart Augmentation Works}

The proposed method uses a network (network A) to learn the best sample blending for the specific problem. The output of network A is the used for the input of network B. The idea is to use network A to learn the best data augmentation to train network B.
Network A accepts several samples from the same class in the dataset and generates a new sample from that class, and this new sample should reduce the training loss for network B. 
In figure 3 we see an output of network A designed to do the gender classification. The image on the left is a merged image of the other two. This image represents a sample from the class ``male'' that does not appear in the dataset, but still, has the identifying features of its class.  

\begin{figure}[h]
  \includegraphics[width=8cm]{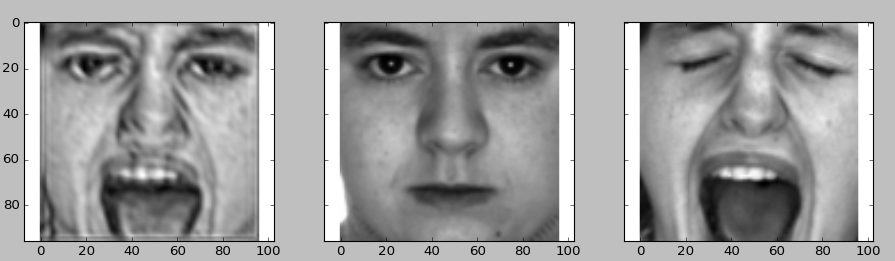}
  \caption{ The image on the left is created by a learned combination of the two images on the right. This type of image transformation helped increase the accuracy of network B. The image was not produced to be an ideal approximation of a face but instead, contains features that helped network B better generalize the concept of gender which is the task it was trained for.  }
\end{figure}

Notice that in figure 3, an image was created with an open mouth and open eyes from two images.  The quality of the face image produced by network A does not matter. Only its ability to help network B better generalize. 
Our approach is most applicable to classification tasks but may also have applications in any approach where the selective blending of sample features improves performance. Our observations show that this approach can reduce overfitting and increase accuracy. 
In the following sections, we evaluate several implementations of our smart augmentation technique on various datasets to show how it can improve accuracy and prevent overfitting. We also show that with smart augmentation, we can train a very small network to perform as well as (or better than) a much larger network that produces state of the art results. 

\section{Methods}

Experiments were conducted on NVIDIA Titan X GPU's running a pascal architecture with python 2.7, using the Theano \cite{2016arXiv160502688short} and Lasange frameworks. 

\subsection{Data Preparation}

To evaluate our method, we chose 4 datasets with characteristics that would allow us to examine the performance of the algorithm on specific types of data. 
Since the goal of our paper is to measure the impact of the proposed technique, we do not attempt to provide a comparison of techniques that work well on these databases. For such a comparison we refer to \cite{MAICS2016} for gender datasets or \cite{zhou2014learning} for the places dataset. 

\subsubsection{Highly constrained faces dataset (db1)}
Our first dataset, db1 was composed from the AR faces database \cite{arface} with a total of 4,000 frontal faces of male and female subjects. The data was split into subject exclusive training, validation, and testing sets, with 70\% for training, 20\% for validation, and 10\% for testing. All face images were reduced to 96X96 grayscale with pixel values normalized between 0 and 1. 
\subsubsection{Augmented, highly constrained faces dataset (db1a)}
To compare traditional augmentation with smart augmentation and to examine the effect of traditional augmentation on smart augmentation, we created an augmented version of db1 with every combination of flipping, blurring, and rotation (-5,-2,0,2,5 degrees with the axis of rotation at the center of the image). This resulted in a larger training set of 48360 images. The test and validation sets were unaltered from db1. The data was split into a subject exclusive training, validation, and testing sets with 70\% for training, 20\% for validation, and 10\% for testing. All face images were reduced to 96X96 with pixel values normalized between 0 and 1. 

\subsubsection{FERET}
Our second dataset, db2, was the FERET dataset. We converted FERET to grayscale and reduced the size of each image to 100X100 with pixel values normalized between 0 and 1. The data was split into subject exclusive training, validation, and testing sets, with 70\% for training, 20\% for validation and 10\% for testing.

\begin{figure}[ht!]
  \includegraphics[width=2cm]{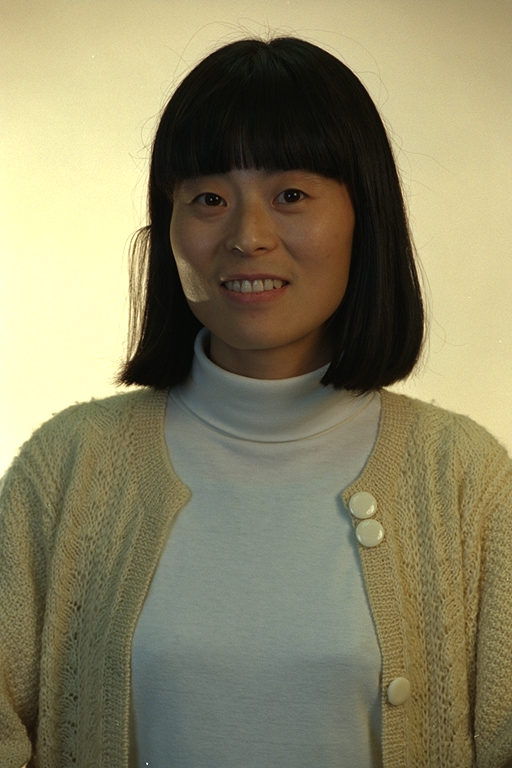}
  \includegraphics[width=2cm]{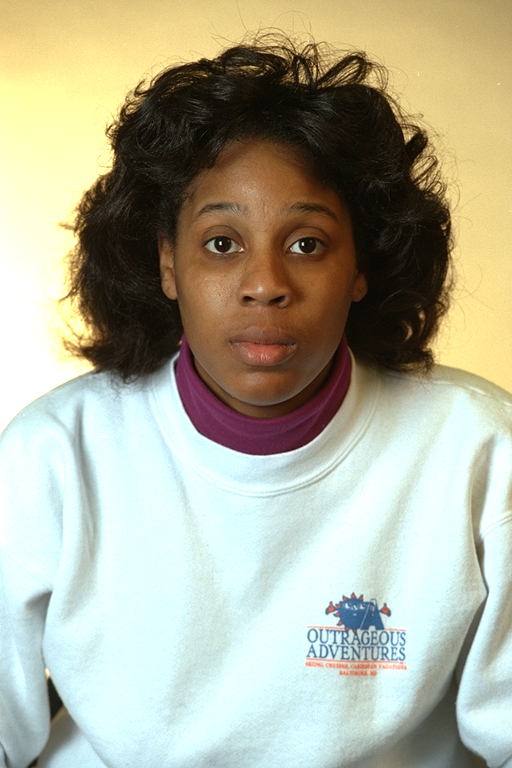}
  \includegraphics[width=2cm]{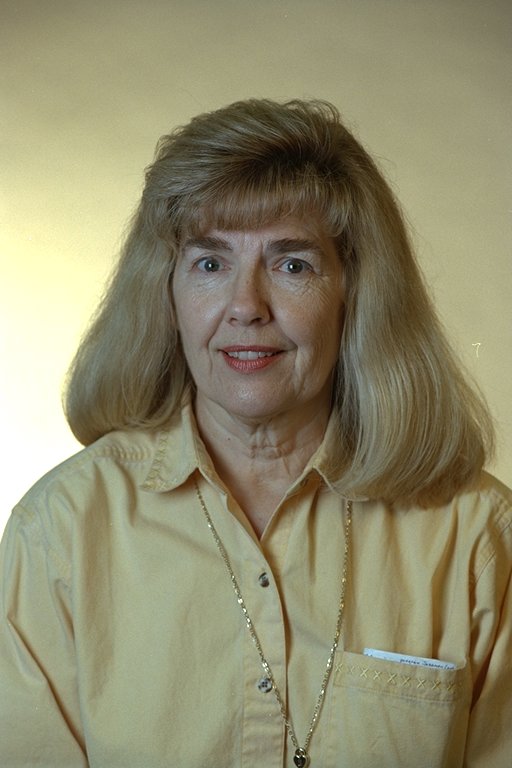}
  \includegraphics[width=2cm]{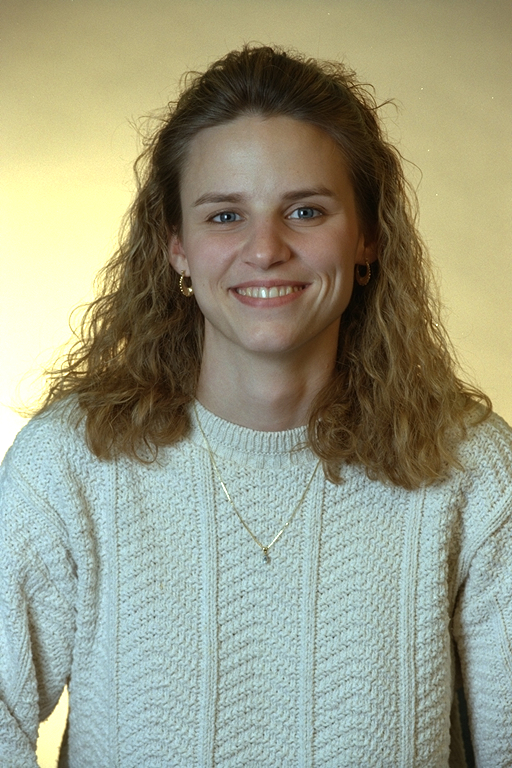}
  \caption{Arbitrarily selected images from FERET  demonstrate similarities in lighting, pose, subject, background, and other photographic conditions. }
\end{figure}

Color FERET \cite{phillips2000feret} Version 2 was collected between December 1993 and August 1996 and made freely available with the intent of promoting the development of face recognition algorithms.  The images are labeled with gender, pose, and name. 

Although FERET contains a large number of high-quality images in different poses and with varying face obstructions (beards, glasses, etc), they all have certain similarities in quality, background, pose, and lighting that make them very easy for modern machine learning methods to correctly classify. In our experiments, we use all images in FERET for which gender labels exist. 

\subsubsection{Adience}
Our third dataset, db3, was Adience. We converted Adience to grayscale images with size 100x100 and normalized the pixel values between 0 and 1. The data was split into subject exclusive training, validation, and testing sets, with 70\% for training, 20\% for validation and 10\% for testing.

\begin{figure}[ht!]
  \includegraphics[width=2cm]{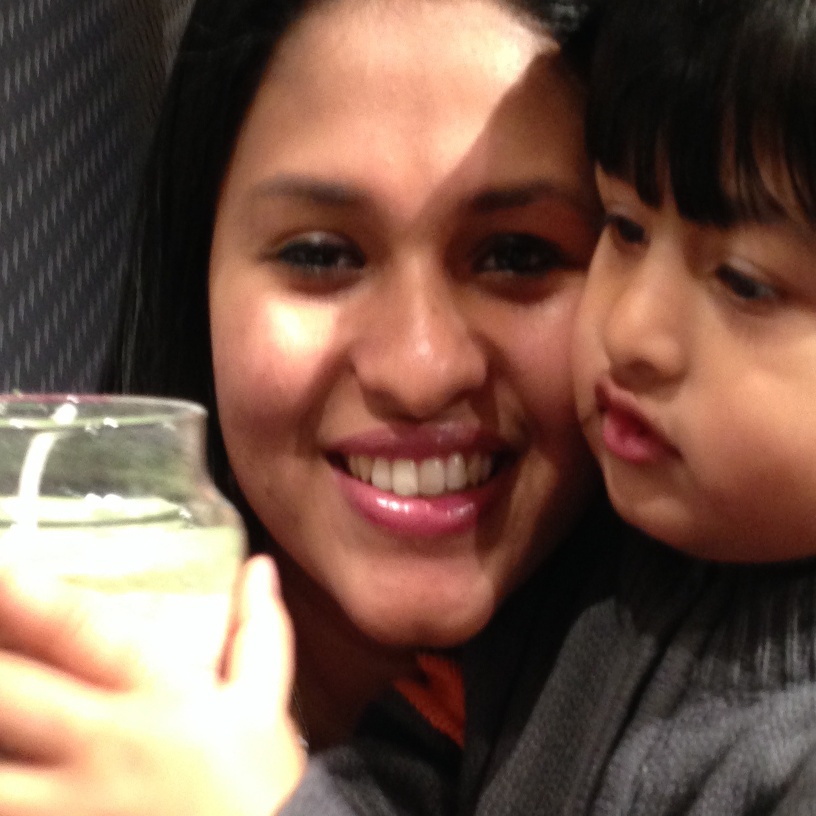}
  \includegraphics[width=2cm]{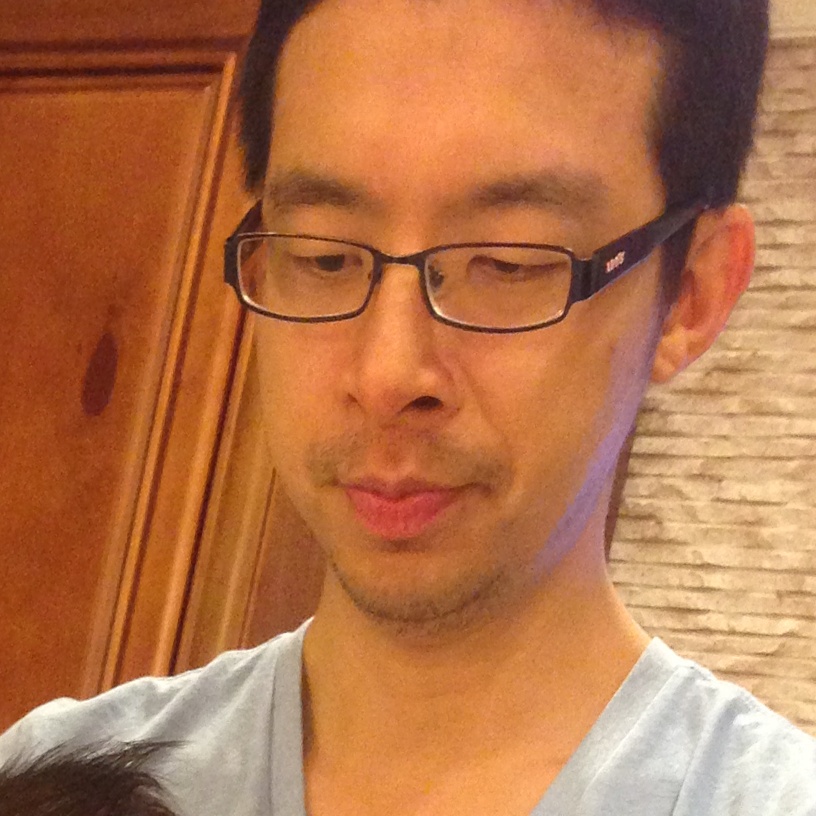}
  \includegraphics[width=2cm]{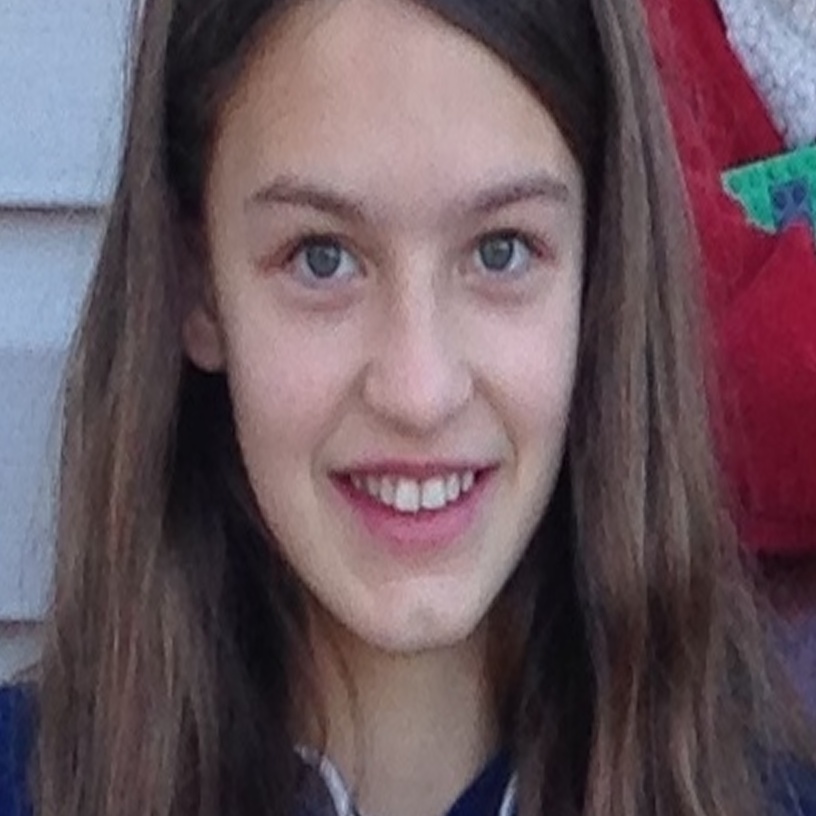}
  \includegraphics[width=2cm]{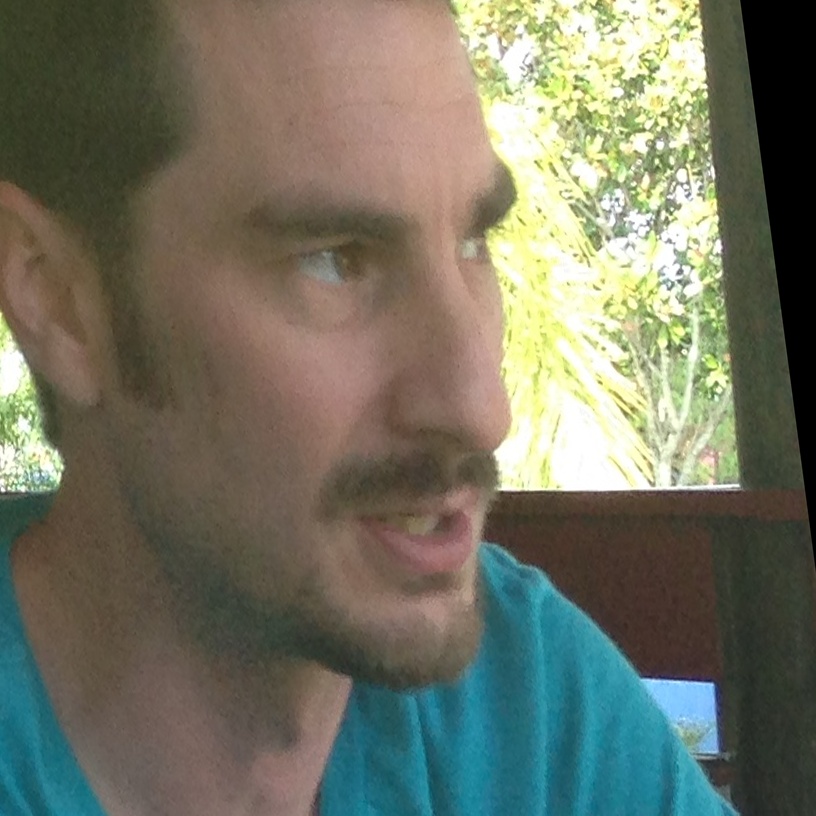}
  \caption{Arbitrarily selected images from the Adience  show significant variations in lighting, pose, subject, background, and other photographic conditions. }
\end{figure}

\subsubsection{DB4}

Our fourth dataset, db4, was the MIT places dataset \cite{zhou2014learning}. The MIT PLACES dataset is a machine learning database containing has 205 scene categories and 2.5 million labeled images.

The Places Dataset is unconstrained and includes complex scenery in a variety of lighting conditions and environments. 

We restricted ourselves to just the first two classes in the dataset (Abbey and Airport). Pixel values were normalized between 0 and 1. The "small dataset," which had been rescaled to 256x256 with 3 color channels, was used for all experiments without modification except for normalization of the pixel values between 0 and 1. 

\begin{figure}[ht!]
  \includegraphics[width=2cm]{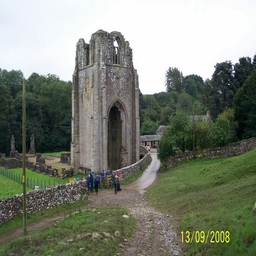}
  \includegraphics[width=2cm]{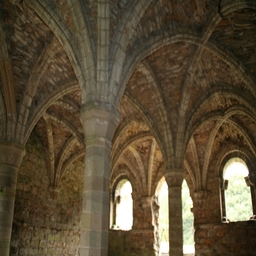}
  \includegraphics[width=2cm]{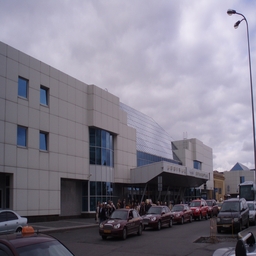}
  \includegraphics[width=2cm]{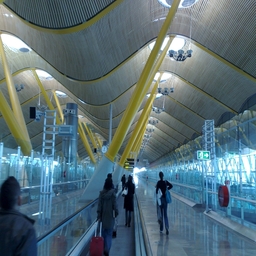}
  \caption{Example images from the MIT places dataset showing two examples from each of the two classes (abbey and airport) used in our experiments. }
\end{figure}

\section{Experiments}

In these experiments, we call network B the network that is being trained for a specific task (such as classification). We call network A the network that learns augmentations that help train network B. 

All experiments are run for 1000 epochs. The test accuracy reported is for the network that had the highest score on the validation set during those 1000 epochs.  

To analyze the effectiveness of Smart Augmentation, we performed 30 experiments using 4 datasets with different parameters. A brief overview of the experiments can be seen in Table I. The experiments were conducted with the motivation of answering the following questions:

\begin{enumerate}

\item Is there any difference in accuracy between using smart augmentation and not using it? (Is smart augmentation effective?)
\item If smart augmentation is effective, is it effective on a variety of datasets? 
\item As the datasets become increasingly unconstrained, does smart augmentation perform better or worse?
\item What is the effect of increasing the number of channels in the smart augmentation method?
\item Can smart augmentation improve accuracy over traditional augmentation? 
\item If smart augmentation and traditional augmentation are combined, are the results better or worse than not combining them?
\item Does altering the $\alpha$ and $\beta$ parameters change the results?
\item Does Smart Augmentation increase or decrease overfitting as measured by train/test loss ratios? 
\item If smart augmentation decreases overfitting, can we use it to replace a large complex network with a simpler one without losing accuracy?
\item What is the effect of the number of network A's on the accuracy? Does training separate networks for each class improve the results?
\end{enumerate}

As listed below, we used three neural network architectures with varied parameters and connection mechanisms. In our experiments, these architectures were combined in various ways as specified in table \ref{fullexperiments}.

\begin{itemize}
\item Network $B_1$  is a simple, small Convolutional neural network, trained as a classifier, that takes an image as input, and outputs class labels with a softmax layer. This network is illustrated in figure 7.
\item Network $B_2$ is a unmodified implementation of VGG16 as described in \cite{simonyan2014very}. Network $B_2$ is a large network that takes an image as input and outputs class labels with a softmax layer. 
\item Network $A$ is a Convolutional neural network that takes one or more images as input and outputs a modified image. 

\end{itemize}

\begin{figure}[h]
  \includegraphics[width=9cm]{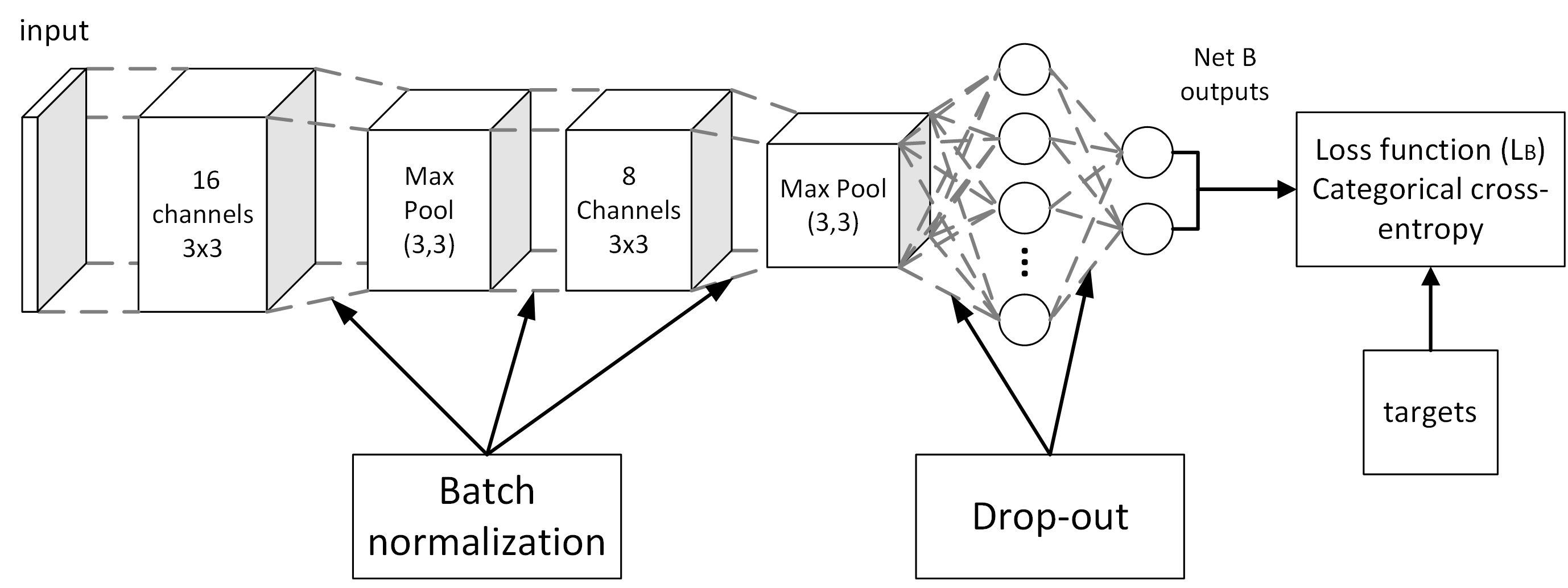}
  \caption{Illustration of network $B_1$}
\end{figure}

\begin{figure}[h]
  \includegraphics[width=9cm]{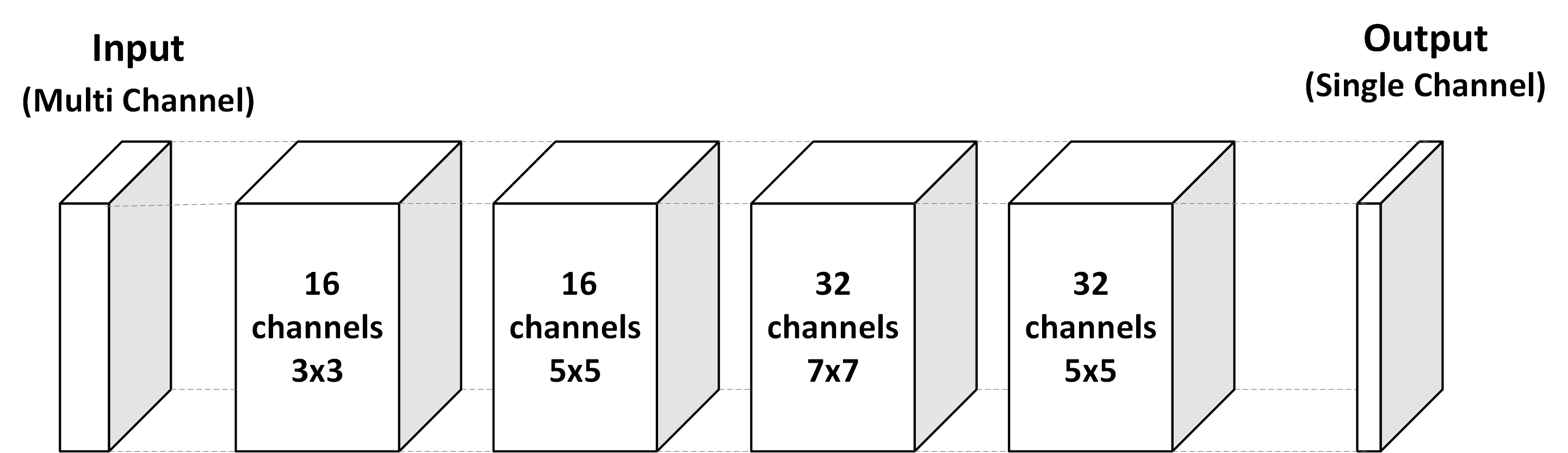}
  \caption{Illustration of network $A$}
\end{figure}

\begin{table}[]
\centering
\caption{Full listing of experiments.}
\label{fullexperiments}
\begin{tabular}{|c|c|c|c|c|c|c|c|c|}

\hline
Exp  & DB     &Net A & A ch & Net B & $\alpha$   & $\beta$  & LR & Momentum \\ \hline
1       & db1    & 1         & 1          & B\_1  & 0.3 & 0.7 & 0.01  & 0.9      \\ \hline
2       & db1    & 1         & 2          & B\_1  & 0.3 & 0.7 & 0.01  & 0.9      \\ \hline
3       & db1    & 1         & 3          & B\_1  & 0.3 & 0.7 & 0.01  & 0.9      \\ \hline
4       & db1    & 1         & 4          & B\_1  & 0.3 & 0.7 & 0.01  & 0.9      \\ \hline
5       & db1    & 1         & 5          & B\_1  & 0.3 & 0.7 & 0.01  & 0.9      \\ \hline
6       & db1    & 1         & 6          & B\_1  & 0.3 & 0.7 & 0.01  & 0.9      \\ \hline
7       & db1    & 1         & 7          & B\_1  & 0.3 & 0.7 & 0.01  & 0.9      \\ \hline
8       & db1    & 1         & 8          & B\_1  & 0.3 & 0.7 & 0.01  & 0.9      \\ \hline
9       & db1    & 2         & 1          & B\_1  & 0.3 & 0.7 & 0.005 & 0.9      \\ \hline
10      & db1    & 2         & 2          & B\_1  & 0.3 & 0.7 & 0.005 & 0.9      \\ \hline
11      & db1    & 2         & 3          & B\_1  & 0.3 & 0.7 & 0.005 & 0.9      \\ \hline
12      & db1    & 2         & 4          & B\_1  & 0.3 & 0.7 & 0.005 & 0.9      \\ \hline
13      & db1    & 2         & 5          & B\_1  & 0.3 & 0.7 & 0.005 & 0.9      \\ \hline
14      & db1    & 2         & 6          & B\_1  & 0.3 & 0.7 & 0.005 & 0.9      \\ \hline
15      & db1    & 2         & 7          & B\_1  & 0.3 & 0.7 & 0.005 & 0.9      \\ \hline
16      & db1    & 2         & 8          & B\_1  & 0.3 & 0.7 & 0.005 & 0.9      \\ \hline
17      & db1    & NA        & NA         & B\_1  & NA  & NA  & 0.01  & 0.9      \\ \hline
18      & db1a & NA        & NA         & B\_1  & NA  & NA  & 0.01  & 0.9      \\ \hline
19      & db1a & 1         & 2          & B\_1  & 0.3 & 0.7 & 0.01  & 0.9      \\ \hline
20      & db1a & 2         & 2          & B\_1  & 0.3 & 0.7 & 0.005 & 0.9      \\ \hline
21      & db2    & NA        & NA         & B\_1  & NA  & NA  & 0.01  & 0.9      \\ \hline
22      & db2    & 1         & 2          & B\_1  & 0.3 & 0.7 & 0.01  & 0.9      \\ \hline
23      & db3    & NA         & NA         & B\_1  & NA  & NA  & 0.01  & 0.9      \\ \hline
24      & db3    & 1         & 2          & B\_1  & 0.3 & 0.7 & 0.01  & 0.9      \\ \hline
25      & db4    & NA        & NA         & B\_2  & NA  & NA  & 0.005 & 0.9      \\ \hline
26      & db4    & NA        & NA         & B\_1  & NA  & NA  & 0.005 & 0.9      \\ \hline
27      & db4    & 1         & 2          & B\_1  & 0.3 & 0.7 & 0.01  & 0.9      \\ \hline
28      & db4    & 1         & 2          & B\_1  & 0.7 & 0.3 & 0.01  & 0.9      \\ \hline
29      & db4    & 2         & 2          & B\_1  & 0.7 & 0.3 & 0.005 & 0.9      \\ \hline
30      & db4    & 2         & 2          & B\_1  & 0.3 & 0.7 & 0.005 & 0.9      \\ \hline
\end{tabular}
\end{table}

\subsection{Smart Augmentation with one Network A on the Gender Classification Task}

Experiments 1-8, 19,22, and 24  as seen in table \ref{fullexperiments} were trained for gender classification using the same technique as illustrated in figure 9. In these experiments, we use smart augmentation to train a network (network B) for gender classification using the specified database. 

\begin{figure*}[h]
  \includegraphics[width=\textwidth]{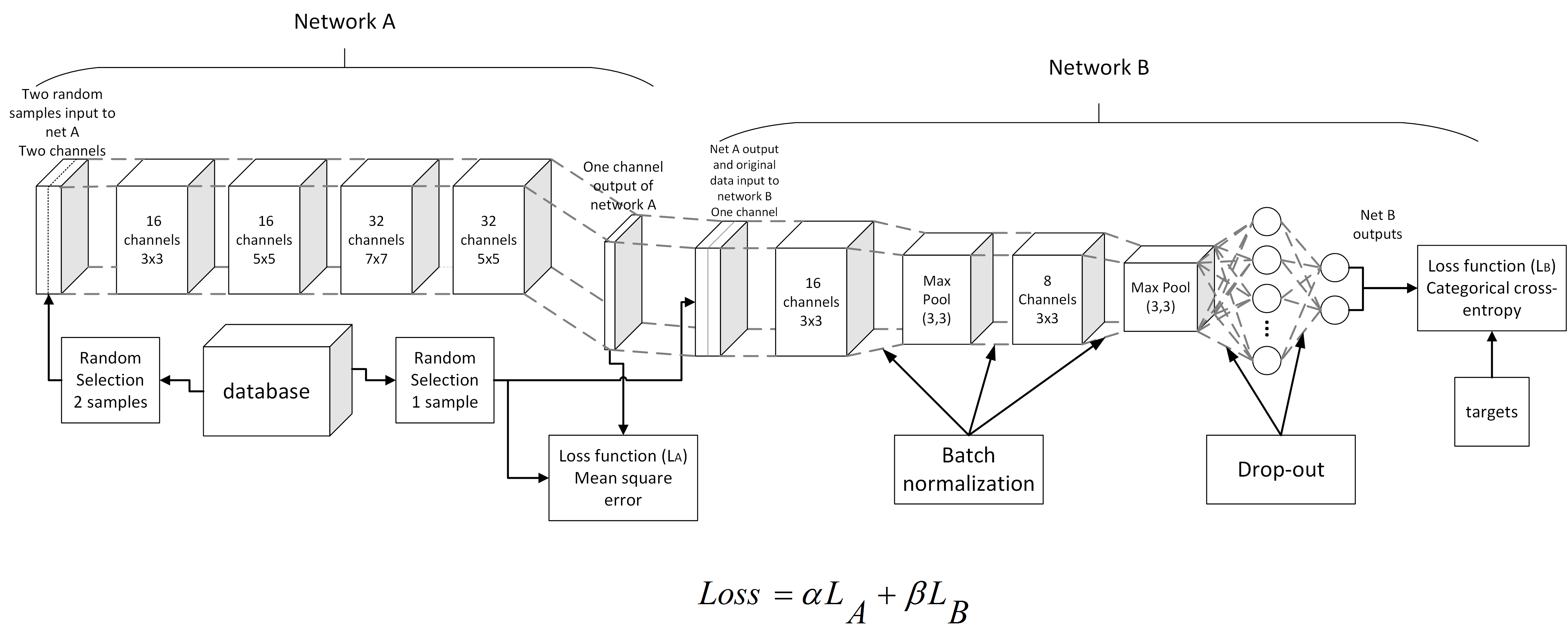}
  \caption{Diagram of simplified implementation of Smart Augmentation showing network A and network B}
\end{figure*}

The first, $k$ images are randomly selected from the same class (male or female) in the dataset. These $k$ samples are merged into $k$ channels of a single sample. The grayscale values of the first image, $img_0$, are mapped to channel 0 and the grayscale values of the second image $im_1$ are mapped to channel 1 and so on until we reach the number of channels specified in the experiments table.  This new $k$ channel image is fed into the network A. Network A is a fully convolutional neural network (See figure 8) which accepts images as the input and gives the images with the same size at the output in a single channel. 

An additional grayscale image is then randomly selected from the same class in the dataset (this image should not be any of those images selected in step 1). The loss function for this network A is calculated as the mean squared error between this randomly selected image and the output of network A. The output of network A, and the target image is then fed into network B as separate inputs. Network B is a typical deep neural network with two convolutional layers followed by batch normalization and max-pooling steps after each convolutional layer. Two fully connected layers are placed at the end of the network. The first of these layers has 1024 units and the second dense layer is made of two units as the output of network B using softmax. Each dense layer takes advantage of the drop-out technique in order to avoid over-fitting. The loss function of network B is calculated as the categorical cross-entropy between the outputs and the targets.

The total loss of the whole model is a linear combination of the loss functions of two networks. This approach is designed to train a network A that generates samples which reduce the error for network B. The validation loss was calculated only for network B, without considering network. This allows us to compare validation loss with and without smart augmentation. 

Our models were trained using Stochastic Gradient Descent with Nesterov Momentum \cite{sutskever2013importance}, learning rate 0.01 and momentum 0.9. The lasagne library used to train the network in python. 

In these experiments, we varied the number of input channels and datasets used. Specifically, we trained a network B from scratch with 1-8 input channels with a single network A on db1, 2 channels on network A for db2 and 3, and 2 channels on network db1a as shown in the table of experiments. 

\subsection{Smart Augmentation with two Network A's on the Gender Classification Task}

In experiments 9-16 and 20 we evaluate a different implementation of smart augmentation, containing a separate network A for each class. As before, the first $k$ images are randomly selected from the same class (male or female) in the dataset. These $k$ samples are merged into $k$ channels of a single sample.The grayscale values of the first image, $img_0$, are mapped to channel 0 and the grayscale values of the second image, $im_1$, are mapped to channel 1, and so on until we reach the number of channels specified in the experiments table just as before. Since we now have two network A's, it is important to separate out the loss functions for each network as illustrated in figure 9.

All other loss functions are calculated the same way as before. 

One very important difference is the updated learning rate (0.005). While performing initial experiments we noticed that using a learning rate above 0.005 led to the ``dying RELU'' problem and stopped effective learning within the first two epochs. This network is also more sensitive to variations in batch size. 

The goal of these experiments was to examine how using multiple network A’s impacts accuracy and overfitting compared to just using one network A. We also wanted to know if there were any differences when trained on a manually augmented database (experiment 20).  

\begin{figure*}[h]
  \includegraphics[width=\textwidth]{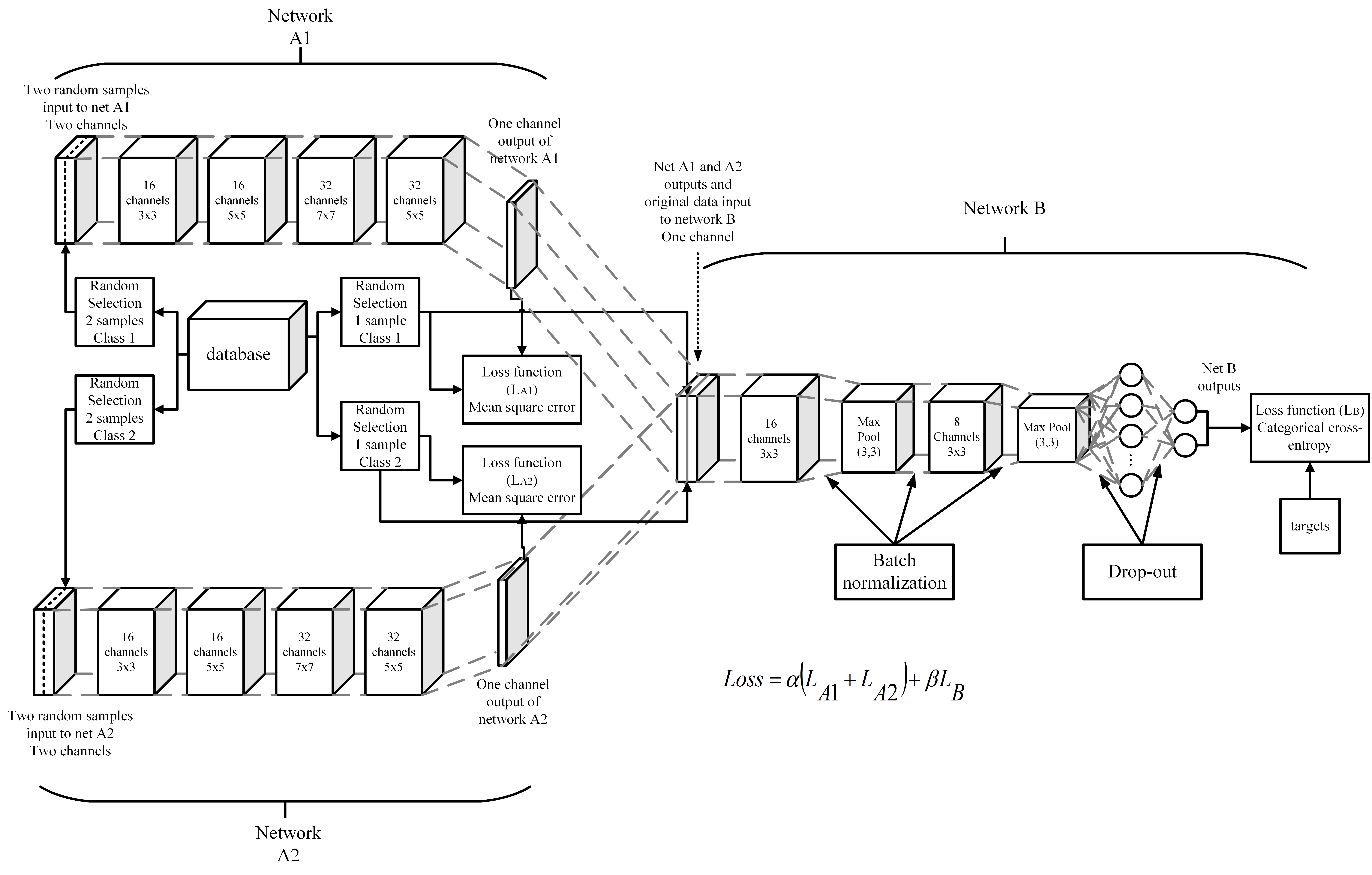}
  \caption{Diagram of our implementation of Smart Augmentation with one network A for each class}
\end{figure*}

\subsection{Training without Smart Augmentation on the Gender Classification Task}

In these experiments, we train a network (network B) to perform gender classification without applying network A during the training stage. These experiments (23, 21, 18, and 17)  are intended to serve as a baseline comparison of what network B can learn without smart augmentation on a specific dataset (db3,db2, db1a, and db1 respectively). In this way, we measure any improvement given by smart augmentation. A full implementation of Network B is shown in figure 7.

\begin{figure}[h]
  \includegraphics[width=9cm]{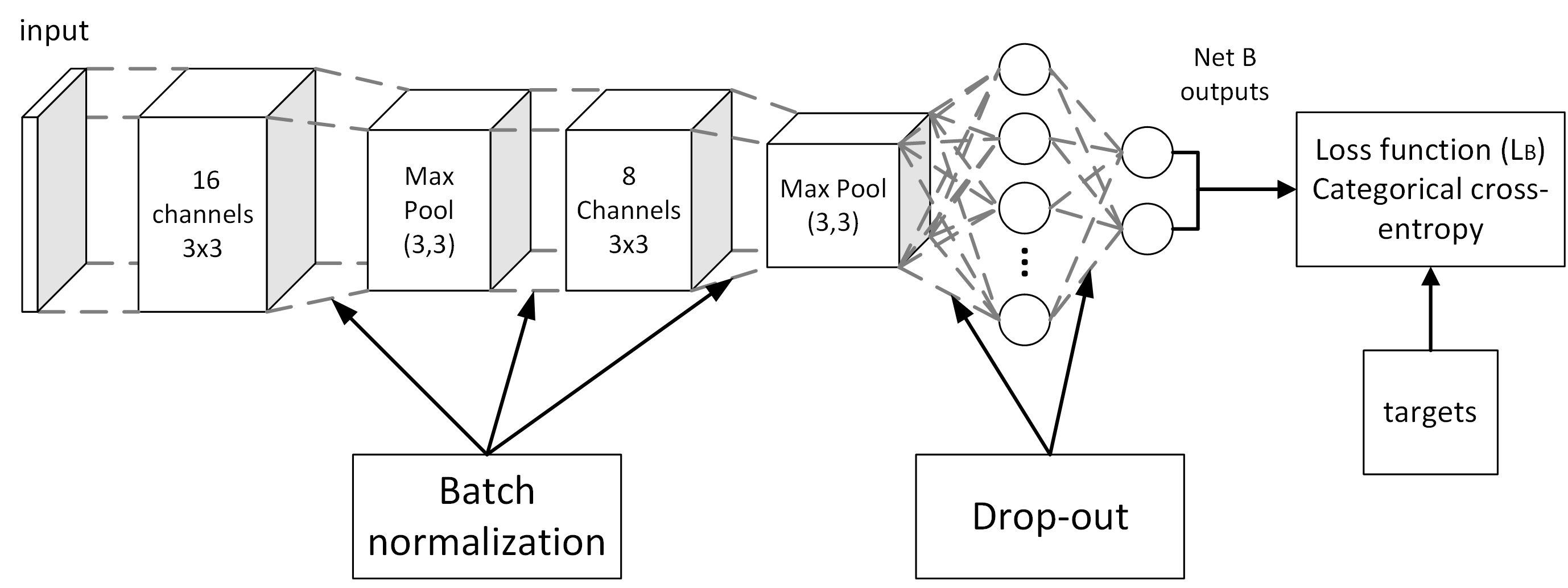}
  \caption{Diagram of implementation of network B without Smart Augmentation}
\end{figure}

This network has the same architecture as the network B presented in the previous experiment except that it does not utilize a network A. 

As before, two fully connected layers are placed at the end of the network. The first of these layers has 1024 units, and the second dense layer has two units (one for each class). Each dense layer takes advantage of the drop-out technique in order to avoid over-fitting. 

All loss functions (training, validation, and testing loss) were calculated as the categorical cross-entropy between the outputs and the targets.

As before, models were trained using Stochastic Gradient Descent with Nesterov Momentum \cite{sutskever2013importance}, learning rate 0.01 and momentum 0.9. The lasagne library was used to train the network in python. 

\subsection{Experiments on the places dataset}

In the previous experiments in this section, we used 3 different face datasets. In experiments 25 - 30 we examine the suitability of Smart Augmentation with color scenes from around the world from the MIT Places dataset to evaluate our method on data of a completely different topic. We varied the $\alpha$ and $\beta$ parameter in our global loss function so that we could identify how they influence results. Unlike in previous experiments, we also retained color information. 

Experiment 25 utilized a VGG16 trained from scratch as a classifier, chosen because VGG16 models have performed very well on the places dataset in public competitions \cite{zhou2014learning}. The input to network A was 256x256 RGB images and the output was determined by a 2 class softmax classifier. 

In experiment 26 we use a network B, identical in all respects to the one used in the previous subsection, except that we use the lower learning rate specified in the experiments table and take in color images about places instead of gender. 

These two experiments (25,26) involved simple classifiers to establish a baseline against which other experiments on the same dataset could be evaluated. 

In experiments 27-28, $k$ images were randomly selected from the same class (abbey or airport) in the dataset. These $k$ samples are merged into $k \times 3$ channels of a single sample. The values of the first three channels of image $img_0$ are mapped to channel 0-2, and the first three channels of the second image $im_1$ are mapped to channels 3-5, and so on, until we reach the number of channels specified in the experiments table multiplied by the number of color channels in the source images.  This new $k \times 3$ channel image is used by network A. Network A is a fully convolutional neural network) which accepts images as the input, and outputs just one image. 

An additional image is then randomly selected from the same class in the dataset. The loss function for network A is calculated as the mean squared error between the randomly selected image and the output of network A. The output of network A, and the target image is then fed into network B as separate inputs. Network B is a typical deep neural network with two convolutional layers followed by batch normalization and max-pooling steps after each convolutional layer. Two fully connected layers are placed at the end of the network. The first of these layers has 1024 units and the second dense layer is made of two units as the output of network B using softmax. Each dense layer takes advantage of the drop-out technique in order to avoid over-fitting. The loss function of network B is calculated as the categorical cross-entropy between the outputs and the targets.

The total loss of the whole model is a linear combination of the loss functions of two networks. This approach is designed to train a network A that generates samples that reduce the error for network B. The validation loss was calculated only for network B, without considering network A. This allows us to compare validation loss with and without smart augmentation. 

Our models were trained using Stochastic Gradient Descent with Nesterov Momentum \cite{sutskever2013importance}, learning rate 0.005 and momentum 0.9. The Lasagne library was used to train the network in python. 

In these experiments, we varied the number of input channels and datasets used. Specifically, we trained a network B from scratch with 1-8 input channels on network A on db1, 2 channels on network A for db2 and 3, and 2 channels on network db1a as shown in the table of experiments. 

In experiments 29-30, $k$ images are randomly selected from the same class (abbey or airport) in the dataset. These $k$ samples are merged into $k \times 3$ channels of a single sample. The values of the first three channels in image $img_0$ are mapped to channel 0-2 and the first three channels of the second image $im_1$ are mapped to channels 3-5 and so on until we reach the number of channels specified in the experiments table multiplied by the number of color channels in the source images.  This new $k\times 3$ channel image is fed into the network A. Network A is a fully convolutional neural network which accepts images as the input and outputs a single color image. 

An additional image is then randomly selected from the same class in the dataset. The loss function for each network A is calculated as the mean squared error between the randomly selected image and the output of network A. The output of network A, and the target image is then fed into network B as separate inputs. Network B is a typical deep neural network with two convolutional layers followed by batch normalization and max-pooling steps after each convolutional layer. Two fully connected layers are placed at the end of the network. The first of these layers has 1024 units, and the second dense layer is made of two units as the output of network B using softmax. Each dense layer takes advantage of the drop-out technique in order to avoid over-fitting. The loss function of network B is calculated as the categorical cross-entropy between the outputs and the targets.

The total loss of the whole model is a linear combination of the loss functions of the two networks. This approach is designed to train a network A that generates samples that reduce the error for network B. The validation loss was calculated only for network B, without considering network A. This allows us to compare validation loss with and without smart augmentation. 

Our models were trained using Stochastic Gradient Descent with Nesterov Momentum \cite{sutskever2013importance}, learning rate 0.005 and momentum 0.9. The lasagne library was used to train the network in python. 

In these experiments, we varied the number of input channels and datasets used. Specifically, we trained a network B from scratch with 1-8 input channels on network A on db1, 2 channels on network A for db2 and 3, and 2 channels on network db1a as shown in the table of experiments. 

\section{Results}

\begin{table}[]
\centering
\caption{Results of Experiments on Face Datasets}
\label{faceresults}
\begin{tabular}{|c|c|c|c|c|}
\hline
\multicolumn{5}{|c|}{\textbf{Experiments on Face Datasets}}                                                           \\ \hline
\textbf{Dataset} & \textbf{\#Net As} & \textbf{Input Channels} & \textbf{Augmented} & \textbf{Test Accuracy} \\ \hline
AR Faces     & 1                 & 1                       & no                          & 0.927746               \\ \hline
AR Faces     & 1                 & 2                       & no                          & 0.924855               \\ \hline
AR Faces     & 1                 & 3                       & no                          & 0.950867               \\ \hline
AR Faces     & 1                 & 4                       & no                          & 0.916185               \\ \hline
AR Faces    & 1                 & 5                       & no                          & 0.910405               \\ \hline
AR Faces   & 1                 & 6                       & no                          & 0.933526               \\ \hline
AR Faces     & 1                 & 7                       & no                          & 0.916185               \\ \hline
AR Faces     & 1                 & 8                       & no                          & 0.953757               \\ \hline
AR Faces    & 2                 & 1                       & no                          & 0.869942188            \\ \hline
AR Faces     & 2                 & 2                       & no                          & 0.956647396            \\ \hline
AR Faces     & 2                 & 3                       & no                          & 0.942196548            \\ \hline
AR Faces     & 2                 & 4                       & no                          & 0.942196548            \\ \hline
AR Faces     & 2                 & 5                       & no                          & 0.907514453            \\ \hline
AR Faces    & 2                 & 6                       & no                          & 0.933526039            \\ \hline
AR Faces     & 2                 & 7                       & no                          & 0.916184962            \\ \hline
AR Faces     & 2                 & 8                       & no                          & 0.924855471            \\ \hline
AR Faces     & 0                 & NA                      & no                          & 0.881502867            \\ \hline
AR Faces    & 0                 & NA                      & yes                         & 0.890173435            \\ \hline
AR Faces     & 1                 & 2                       & yes                         & 0.956647396            \\ \hline
AR Faces     & 2                 & 2                       & yes                         & 0.956647396            \\ \hline
Adience          & 0                 & NA                      & no                          & 0.700206399            \\ \hline
Adience          & 1                 & 2                       & no                          & 0.760577917            \\ \hline
FERET            & 0                 & NA                      & no                          & 0.835242271            \\ \hline
FERET            & 1                 & 2                       & no                          & 0.884581506            \\ \hline
\end{tabular}
\end{table}

\begin{table}[]
\centering
\caption{Results of Experiments on Place Dataset}
\label{my-label}
\begin{tabular}{|c|c|c|c|c|}
\hline
\multicolumn{5}{|c|}{\textbf{Experiments on MIT places dataset}}                               \\ \hline
\textbf{\#Net As} & \textbf{Target Network} & \textbf{Test Accuracy} & \textbf{A} & \textbf{B} \\ \hline
0                 & VGG16                   & 98.5                   & NA         & NA         \\ \hline
0                 & Small net B             & 96.5                   & NA         & NA         \\ \hline
1                 & Small net B             & 98.75                  & 0.3        & 0.7        \\ \hline
1                 & Small net B             & 99\%                   & 0.7        & 0.3        \\ \hline
2                 & Small net B             & 99\%                   & 0.7        & 0.3        \\ \hline
2                 & Small net B             & 97.87\%                & 0.3        & 0.7        \\ \hline
\end{tabular}
\end{table}

The results of experiments 1-30 as shown in Table I are listed in tables II and III and are listed in the same order as in the corresponding experiments table. These results are explained in detail in the subsections below. 

\subsection{Smart Augmentation with one Network A on the Gender Classification Task}

In figure 12, we show the training and validation loss for experiments 1 and 17. As can be observed, the rate of overfitting was greatly reduced when smart augmentation was used compared to when it was not used. 

Without smart augmentation, network B had an accuracy of 88.15 for the AR faces dataset; for the rest of this subsection, this result is used as a baseline by which other results on that dataset are evaluated. 

\begin{figure*}[h]
  \includegraphics[width=\textwidth]{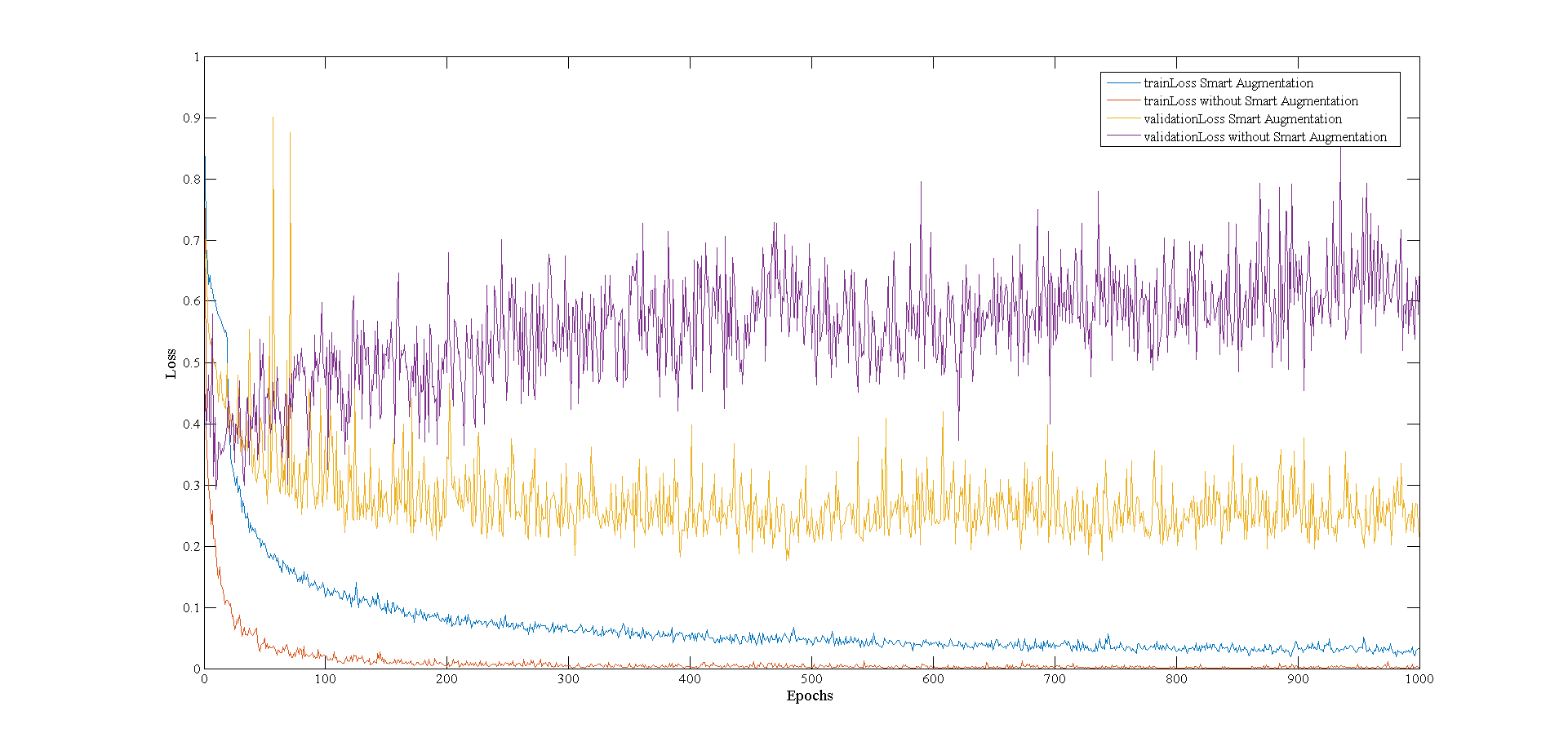}
  \caption{Training and validation losses for experiments 1 and 17, showing reductions in overfitting by using Smart Augmentation.  The smaller difference between training loss and validation loss caused by the smart augmentation technique shows how this approach helps the network B to learn more general features for this task. To avoid confusion, we remind the reader that the loss for smart augmentation is given by $f(L_A, L_B; \alpha, \beta)$. This means that the loss graphs are a combination of the losses of two networks whereas the losses without smart augmentation are only $f( L_B)$.  }
\end{figure*}

One can see how the smart augmentation technique could prevent network B from overfitting in the training stage. The smaller difference between training loss and validation loss caused by the smart augmentation technique shows how this approach helps the network B to learn more general features for this task. 
Network B also had higher accuracy on the test set when trained with smart augmentation.

In figures 13 and 14 we show examples of the kinds of images network A learned to generate. In these figures, the image on the left side is the blended image of the other two images produced by network A.

\begin{figure}[h]
  \includegraphics[width=8cm]{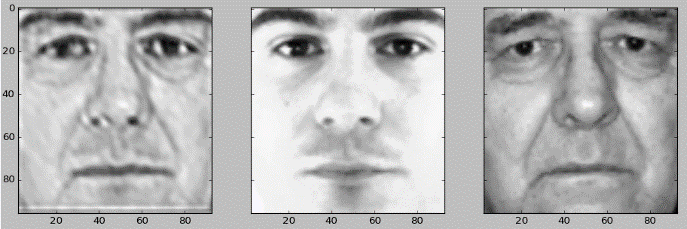}
  \caption{The image on the left is a learned combination of the two images on the right as produced by network A}
\end{figure}

\begin{figure}[h]
  \includegraphics[width=2.666cm]{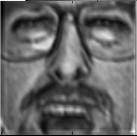}
  \includegraphics[width=2.666cm]{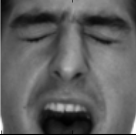}
  \includegraphics[width=2.666cm]{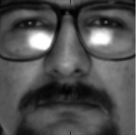}
  \caption{The image on the left is a learned combination of the two images on the right as produced by network A}
\end{figure}

We observe an improvement in accuracy from  83.52\% to  88.46\% from smart augmentation on Feret with 2 inputs and an increase from 70.02\% to 76.06\% on the adience dataset. 

We see that there is no noticeable pattern when we vary the number of inputs for network A. Despite the lack of a pattern, a significant difference was observed with 8 and 3 channels providing the best results at 95.38\%  and  95.09\% respectively. At the lower end, 7, 5, and 4 channels performed the worst, with accuracies of 91.62\%,  91.04\%, and 91.04\%. 

Recall that the accuracy without network A was: 88.15\% for the AR faces dataset.
We suspect that much of the variation in accuracy reported above may be due to chance. Since in this particular experiment, images are chosen randomly there may be times when 2 or more images with very helpful mutual information are present by chance and the opposite is also possible. It is interesting that when 3 and 8 channels were used for network A, the accuracy was over 95\%.

\subsection{Smart Augmentation and Traditional Augmentation}

We note that traditional augmentation improved the accuracy from 88.15\% to 89.08\% without smart augmentation on the gender classification task. When we add smart augmentation we realize an improvement in accuracy to 95.66\% 

The accuracy of the same experiment when we used 2 networks A's was also 95.66\% which seems to indicate that both configurations may have found the same optima when smart augmentation was combined with traditional augmentation. 

This demonstrates that smart augmentation can be used with traditional augmentation to further improve accuracy. In all cases examined so far, Smart Augmentation performed better than traditional augmentation. However, since there are no practical limits on the types of traditional augmentation that can be performed, there is no way to guarantee that manual augmentation could not find a better augmentation strategy. This is not a major concern since we do not claim that smart augmentation should replace traditional augmentation. We only claim that smart augmentation can help with regularization. 

\subsection{Smart Augmentation with two Network A's on the Gender Classification Task}

In this subsection, we discuss the results of our two network architecture when trained on the gender classification set. 

These experiments show that approaches which use a distinct network A for each class, tend to slightly outperform networks with just 1 network A. This seems to provide support for our initial idea that one network A should be used for each class so that class-specific augmentations could be more efficiently learned. If the networks with just 1 and 0 input channels are excluded, we see an average increase in accuracy from 92.94\% to 93.19\% when smart augmentation is used, with the median accuracy going from 92.49\% to 93.35\%. 

There is only one experiment where smart augmentation performed worse than not using smart augmentation. This can be seen in the 9th row of table II where we use only one channel which caused the accuracy to dip to 86.99\%, contrasted with 88.15\% when no smart augmentation is used. This is expected because when only one channel is used, mutual information can not be effectively utilized. This experiment shows the importance of always using at least 2 channels. 

\subsection{Experiments on the Places Dataset}

As with previously discussed results, when the places dataset is used, networks with multiple network A's performed slightly better. We also notice that when $\alpha$ is higher than $\beta$ an increase in accuracy is realized. 

The most significant results of this set of experiments is the comparison between smart augmentation, VGG 16, and network B trained alone. Note that a small network B trained alone (no Smart Augmentation) had an accuracy of 96.5\% compared to VGG 16 (no Smart Augmentation) at 98.5\%. When the same small network B was trained with smart augmentation we see accuracies ranging from  98.75\% to 99\% which indicates that smart augmentation, in some cases, can allow a much smaller network to replace a larger network. 

\section{Discussion and Conclusion}

Smart Augmentation has shown the potential to increase accuracy by demonstrably significant measures on all datasets tested. In addition, it has shown potential to achieve similar or improved performance levels with significantly smaller network sizes in a number of tested cases.

In this paper, we discussed a new regularization approach, called ``Smart Augmentation'' to automatically learn suitable augmentations during the process of training a deep neural network. We focus on learning augmentations that take advantage of the mutual information within a class. The proposed solution was tested on progressively more difficult datasets starting with a highly constrained face database and ending with a highly complex and unconstrained database of places. The various experiments presented in this work demonstrate that our method is appropriate for a wide range of tasks and demonstrates that it is not biased to any particular type of image data. 

As a primary conclusion, these experiments demonstrate that the augmentation process can be automated, specifically in nontrivial cases where two or more samples of a certain class are merged in nonlinear ways resulting in improved generalization of a target network. The results indicate that a deep neural network can be used to learn the augmentation task in this way at the same time the task is being learned. We have demonstrated that smart augmentation can be used to reduce overfitting during the training process and reduce the error during testing. 

It is worthwhile to summarize a number of additional observations and conclusions from the various experiments documented in this research. 

Firstly, no linear correlation between the number of samples mixed by network A and accuracy was found so long as at least 2 samples are used.  

Secondly, it was shown that Smart Augmentation is effective at reducing error and decreasing overfitting and that this is true regardless of how unconstrained the database is. 

Thirdly, these experiments demonstrated that better accuracy could be achieved with smart augmentation than with traditional augmentation alone. It was found that altering the $\alpha$ and $\beta$ parameters of the loss function slightly impacts results but more experiments are needed to identify if optimal parameters can be found. 

Finally, it was found that Smart Augmentation on a small network achieved better results than those obtained by a much larger network (VGG 16). This will help enable more practical implementations of CNN networks for use in embedded systems and consumer devices where the large size of these networks can limit their usefulness. 

Future work may include expanding Smart Augmentation to learn more sophisticated augmentation strategies and performing experiments on larger datasets with larger numbers of data classes. A statistical study to identify the number of channels that give the highest probability of obtaining optimal results could also be useful.  

\section{Acknowledgements}
This research is funded under the SFI Strategic Partnership Program by Science Foundation Ireland (SFI) and FotoNation Ltd. Project ID: 13/SPP/I2868 on Next Generation Imaging for Smartphone and Embedded Platforms. 
This work is also supported by an Irish Research Council Employment Based Programme Award. Project ID: EBPPG/2016/280

We gratefully acknowledge the support of NVIDIA Corporation with the donation of a Titan X GPU used for this research.
\newpage


\IEEEtriggeratref{12}



%

\bibliographystyle{IEEEtran}
\bibliography{smartaugmentation}

\end{document}